\documentclass{article}
\usepackage[preprint]{spconf}
\usepackage{amsmath,graphicx}

\usepackage[hyphens]{url}
\usepackage{hyperref}
\usepackage[hyphenbreaks]{breakurl}
\usepackage{xcolor}
\usepackage{algpseudocode}
\usepackage[ruled,vlined,noend,linesnumbered]{algorithm2e}

\SetCommentSty{mycommfont}

\usepackage{tikz}

\copyrightnotice{%
\begin{tikzpicture}[remember picture,overlay]
\node[anchor=south,yshift=10pt] at (current page.south) {\fbox{\parbox{\dimexpr\textwidth-\fboxsep-\fboxrule\relax}{\footnotesize \copyright\ 2021 IEEE. Personal use of this material is permitted. Permission from IEEE must be obtained for all other uses, in any current or future media, including reprinting/republishing this material for advertising or promotional purposes, creating new collective works, for resale or redistribution to servers or lists, or reuse of any copyrighted component of this work in other works.}}};
\end{tikzpicture}%
}

\title{AN IMPROVEMENT OF OBJECT DETECTION PERFORMANCE USING MULTI-STEP MACHINE LEARNINGS}
\name{Tomoe Kishimoto$^{1}$, Masahiko Saito$^{1}$, Junichi Tanaka$^{1,2}$, Yutaro Iiyama$^{1}$, Ryu Sawada$^{1}$ and Koji Terashi$^{1}$} 
\address{$^{1}$ International Center for Elementary Particle Physics, The University of Tokyo, Tokyo, Japan\\
$^{2}$ Institute for AI and Beyond, The University of Tokyo, Tokyo, Japan}
\begin{document}
%
\maketitle
\begin{abstract}
Connecting multiple machine learning models into a pipeline is effective for handling complex problems. By breaking down the problem into steps, each tackled by a specific component model of the pipeline, the overall solution can be made accurate and explainable. This paper describes an enhancement of object detection based on this multi-step concept, where a post-processing step called the calibration model is introduced. The calibration model consists of a convolutional neural network, and utilizes rich contextual information based on the domain knowledge of the input. Improvements of object detection performance by 0.8--1.9 in average precision metric over existing object detectors have been observed using the new model.
\end{abstract}
\section{Introduction}
\label{sec:intro}

\begin{figure}[htb]
  \centerline{\includegraphics[width=8cm]{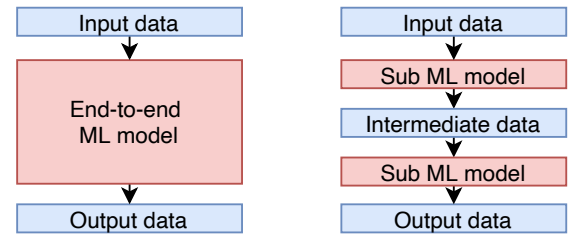}}
  \caption{(a) End-to-end and (b) multi-step ML models.}
  \label{fig:multi-step}
\end{figure}

Machine Learning (ML) has rapidly evolved due to the availability of huge computing power and big data, and has been proven to be successful in many applications such as image classification, natural language translation, etc. 
When a ML solution to a complex problem is seen as a pipeline,
two approaches can be considered as illustrated in Fig.~\ref{fig:multi-step}. If it is possible to build an end-to-end ML model as shown in Fig.~\ref{fig:multi-step}~(a), that is an effective approach in terms of accuracy or speed. However, in many cases, the construction of an end-to-end ML model is not easy. Furthermore, explainability of the result is required in many fields such as medical, scientific, and so on. Therefore, defining multiple ML models step by step as shown in Fig.~\ref{fig:multi-step}~(b) is still a valid approach with merits of:
\begin{itemize}
    \setlength{\leftskip}{-3mm}
 \item Domain knowledge is easily introduced into component models, and component models can be reused in other problems which involve common tasks,
 \item Intermediate data provide the information to understand the behavior of the ML models, which can lead to an explainability of ML. 
\end{itemize}
These two approaches are complementary to each other. In a development cycle of a pipeline, a new idea based on domain knowledge can be examined by the multi-step ML model, which is later replaced by an 
end-to-end model to improve the accuracy and speed.

In this paper, an improvement of object detection performance based on the multi-step ML model concept is reported. Object detection is one of the most fundamental problems in the field of computer vision~\cite{Zou2019ObjectDI}. The main idea for the improvement is to introduce a post-processing ML model that calibrates the outputs of object detectors using contextual information. 
Contextual information, such as the scene an object is placed in, is known to be a vital input for human cognition~\cite{article}, and is therefore expected to enhance object detection capabilities of machine learning models as well. The proposed model uses convolutional neural network (CNN) to effectively handle such information.


This paper is organized as follows. Section~2 describes related works. Section~3 provides details of the proposed calibration model. Section~4 gives results of experiments. Section~5 is the conclusion.

\begin{figure*}[h]
   \centerline{\includegraphics[width=17cm]{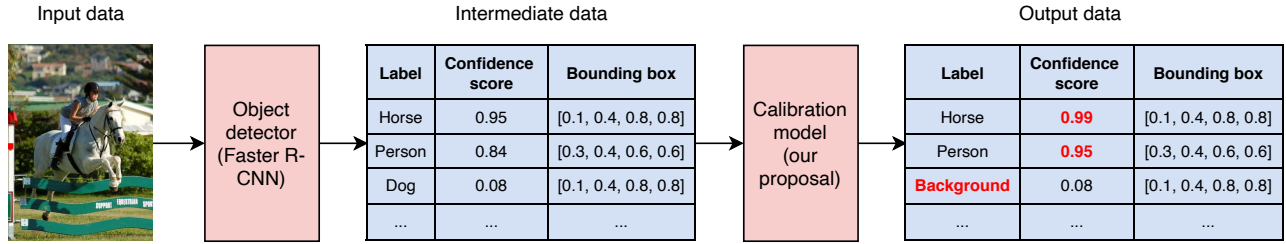}}
  \caption{Schema of multi-step ML models for the object detection.}
  \label{fig:calibration}
\end{figure*}

\section{Related work}
Object detection aims to detect square areas (bounding boxes) in a given image that surround the objects and identify their classes. R-CNN~\cite{6909475} is an object detection model using CNN. R-CNN proposes regions of interest (RoI) using selective search, e.g.~\cite{Uijlings13}, and then extracts CNN features from the RoI. The model is trained to predict the object class and parameters of the bounding box using the CNN features. 
In R-CNN and Fast R-CNN~\cite{7410526}, which is an improvement over the former in terms of speed and accuracy, the RoI proposal relies on a module separated from the network using CNN. Faster R-CNN~\cite{10.5555/2969239.2969250} was proposed to detect the RoI by introducing a region proposal network that shares the full-image CNN features with the detection network in Fast R-CNN. 
While all three models take a two-stage approach separating the RoI proposal and classification, alternative models, such as 
SSD~\cite{DBLP:conf/eccv/LiuAESRFB16}, YOLO~\cite{7780460} and RetinaNet~\cite{journals/corr/abs-1708-02002}, perform object detection in a single step. In this paper, we use Faster R-CNN with ResNet~\cite{7780459} / ResNeXt~\cite{8100117} + FPN~\cite{8099589} backbone model as the baseline detector to evaluate the proposed ML model.

Object detection models using CNN described above do not fully utilize the contextual information, such as co-occurrence of objects, since they mainly focus on individual RoIs. However, importance of such contextual information is well recognized in the computer vision community and its use has been widely studied, e.g.~\cite{7952437, 9137382, sasaki, Chen_2018_ECCV}. 
In contrast to these previous works, our proposed method utilizes rich feature variables, and employs CNN to overcome the problem of ordering of the input variables.

\section{Proposed calibration model}
We now describe a post-processing ML model called calibration model. Figure~\ref{fig:calibration} shows an overview of the multi-step scheme for the object detection.
Faster R-CNN implemented in Detectron2~\cite{wu2019detectron2} is used as the baseline object detector. This baseline detector outputs the class label, confidence score, and bounding box for each detected object. The calibration model uses feature variables which are  calculated from the object detector outputs, and provide new predicted class labels and confidence scores. Details of the calibration model are described in the following. 
\subsection{Feature variables}
\begin{figure}[htb]
        \vspace{-1.8mm}
   \centerline{\includegraphics[width=8 cm]{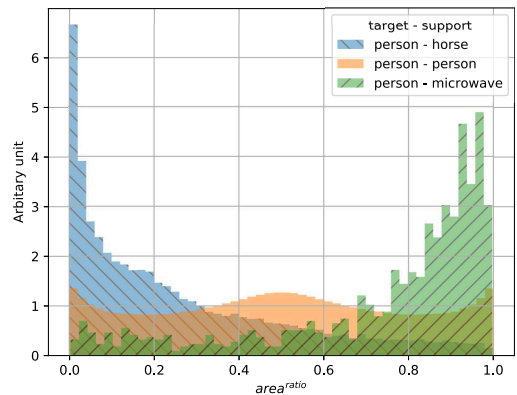}}
 \caption{Distributions of $area^{\rm ratio}$ for selected classes: person - horse, person - person and person - microwave. $area^{\rm ratio}$ is defined as $area^{\rm target}$ / ($area^{\rm target}$ + $area^{\rm support}$).}
  \label{fig:area}
\end{figure}

Feature variables are defined based on the domain knowledge such as object shape, relation between objects, and  so on. In the following, the $target$ bounding box 
is the bounding box whose class label and confidence score we aim to update,
and the $support$ bounding boxes are those detected around the target box. As an example of a features variable, Figure~\ref{fig:area} shows the ratio of areas between the target bounding box and the support bounding box  for three selected classes of the support bounding box. We observe that the target bounding boxes with person class tend to have small area compared to the support bounding boxes with horse class, while the relation is inverted when the support bounding boxes are classified as micorwave. 
These distributions are quite consistent with common sense, and these feature variables are expected to provide good classification power in the calibration model. 

Table~\ref{tab:variables} summarizes the feature variables used in the calibration model. In order to improve the performance, global and local image information are also used as discussed in Section~4.3. All feature variables are normalized between 0 and 1 before being passed to the calibration model.

\begin{table}[htb]
\tabcolsep = 3.pt
  \caption{Summary of feature variables for the calibration model. See Eq.~1 for the definition of the intersection over union. The numbers in bracket indicate the total numbers of variables for each target and support bounding box combination. The total number of class labels is 81 including background.}
    \begin{center}

  \begin{tabular}{|c|c|}\hline\hline
  \multicolumn{2}{|c|}{Basic information and relations (357)} \\\hline\hline
    $(x, y)_{\rm min,max,center}^{\rm target,support}$ & \begin{tabular}{c} Positions of target and support\\ bounding boxes (12)\end{tabular} \\\hline
    \begin{tabular}{c}$area^{\rm target,support}$,\\ $area^{\rm ratio}$ \end{tabular} &  \begin{tabular}{c}Areas of target and support\\ boxes and ratio of $area^{\rm target}$\\ and each $area^{\rm support}$ (3)\end{tabular}\\\hline
     $distance, angle$ & \begin{tabular}{c} Distance and angle between\\ $(x,y)_{\rm center}^{\rm target}$ and $(x,y)_{\rm center}^{\rm support}$ (2)\end{tabular}\\\hline
     \begin{tabular}{c}$iou$, \\ $iou^{target,support}$ \end{tabular} & \begin{tabular}{c} Intersection over union\\ between the target and support\\ bounding boxes (3)\end{tabular}\\\hline
      $n^{\rm support}, n^{\rm overlap}$ & \begin{tabular}{c} Number of support bounding\\ boxes and number of\\ overlapping support bounding\\ boxes for each class index (162)\end{tabular}\\\hline

    $score^{\rm target, support}$ & \begin{tabular}{c} Confidence scores of target\\ and support bounding boxes (2) \end{tabular}\\\hline
    $label^{\rm target, support}$ & \begin{tabular}{c} One-hot vectors of class\\ labels of target and support \\bounding boxes (162)\end{tabular}\\\hline
    \begin{tabular}{c}$aspect^{\rm target, support}$ \\ $aspect^{\rm global}$\end{tabular}  & \begin{tabular}{c} Aspect ratios of target and\\ support bounding boxes, and\\ aspect ratio of the global image (3)\end{tabular}\\\hline
        $edge^{\rm target, support}$ & \begin{tabular}{c} Flags if bounding boxes are on\\ (left, right, top, bottom) edges (8) \end{tabular}\\\hline\hline
      \multicolumn{2}{|c|}{Image information} \\\hline\hline
    \begin{tabular}{c}$image^{\rm target},$ \\ $image^{\rm global}$ \end{tabular} & \begin{tabular}{c} Global image and cropped\\ image of target bounding box,\\ image size is 224 $\times$ 224\end{tabular}\\\hline\hline

  \end{tabular}
    \label{tab:variables}
    \end{center}
\vspace{-8mm}
\end{table}

\subsection{Model components}
\begin{figure}[htb]
\vspace{1.mm}
   \centering\includegraphics[width=8.5 cm]{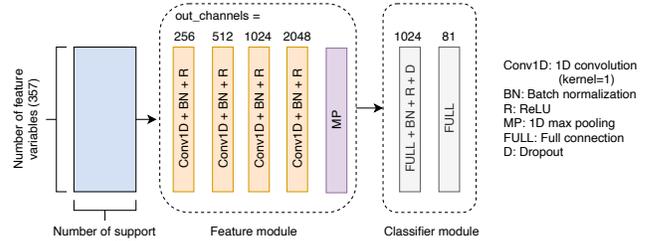}
  \caption{Model description for processing the feature variables.}
  \label{fig:conv1d}
        \vspace{-1.8mm}

\end{figure}

A straightforward method to process the feature variables for the calibration would be to use a Multi-Layer Perceptron (MLP) architecture.
However, MLP models are sensitive to the ordering of the input feature variables for each support box. This is achieved by e.g. ordering the support boxes by $score^{\rm support}$, but we did not observe clear improvement. Therefore, we instead introduced a CNN architecture to the model. Figure.~\ref{fig:conv1d} shows a schematic of the model processing the feature variables.

The input data are prepared to have a two dimensional structure of $M\times N$, where $M$ is the number of support boxes and $N$ is the number of feature variables. The baseline detector is configured to output maximum 100 bounding boxes. Thus, $M$ is 99 after removing the target box. If the baseline detecter outputs less than 100 bounding boxes, the feature variables are zero-padded to obtain a fixed length of $M$. The pointwise convolution (one-dimensional convolution with kernel size = 1) is applied over the features $N$ to extract feature vectors. This one-dimensional convolution is repeated four times with the number of output channels = 256, 512, 1024 and 2048. Finally, the global max-pooling is applied across the support boxes, which allows us to be insensitive to the ordering of  the support boxes. The output vector of the max-pooling, which is 2048 in length, is fed into the classifier to predict the object class. The performance of this calibration model is compared with a simple MLP model, and results are discussed in Section~4.2.

\section{Experiments}

\subsection{Datasets}

Experiments described in this section are based on the COCO 2017 dataset~\cite{10.1007/978-3-319-10602-1_48}, which provides ground truth information of the bounding boxes and their class labels along with real-life images. From each image in the dataset, the baseline detector predicts a list of bounding boxes and their class labels with confidence scores. Only bounding boxes with confidence scores above 0.05 are included in the predictions. These predictions are in turn fed to the calibration model. To perform supervised training of the calibration model, predicted bounding boxes are given truth labels in the following way. For each predicted bounding box, if there exists an overlapping ground truth bounding box with Intersection over Union (IoU) greater than 0.5, the former is given the label of the latter. Otherwise, the predicted bounding box is labeled as background. The IoU is defined as follows:
\begin{equation}
  IoU = \frac{area(R^{p}\cap R^{g})}{area(R^{p}\cup R^{g})},
  \end{equation}
  where $R^{p}$ and $R^{g}$ indicate the predicted and ground truth bounding box, respectively.
Since a large fraction of predicted bounding boxes are labeled as background under this condition, the background is downsampled by a factor 10 in the training dataset to avoid bias. The SGD algorithm with an optimized learning rate is used to train the models. The performance results are obtained from the validation dataset.

\subsection{Classification performance using calibration model}
Table~\ref{tab:class} shows classification performances with different model configurations. In this comparison, ResNet with 50 layers (ResNet-50) is used as the backbone CNN model of the baseline detector. It can be confirmed that the proposed calibration model shows a clear improvement of the f1-score compared to the original baseline detector and the MLP model. 
In the MLP model, the feature module in Fig.~\ref{fig:conv1d} is replaced with 3 full connection layers with 2048 nodes.
In order to archive a further improvement, CNN features, which are extracted from the global image and local image cropped by the target bounding box, are fed into the classifier in Fig.~\ref{fig:conv1d} together with the output of the feature module. ResNet-50 implemented in PyTorch~\cite{NEURIPS2019_9015} is used to extract the CNN features in this comparison. This additional image information leads to f1-score of 0.80. 
In the following, the global and local image information are included in the model when the calibration model is mentioned.

\begin{table}[htb]
          \vspace{-2mm}

  \caption{Summary of classification performance. The values are averages over 81 classes including the background class.}
    \begin{center}

  \begin{tabular}{|cl|c|c|c|}\hline\hline
 & & precision & recall & f1-score\\\hline\hline
(1) & Baseline detector & 0.78& 0.58& 0.66\\\hline
(2) &(1) + MLP model & 0.67& 0.71& 0.69 \\\hline
(3) &(1) + calib. model & 0.75& 0.79& 0.77 \\\hline
(4) & (3) + image info.& 0.78& 0.83& 0.80\\\hline\hline
  \end{tabular}
    \label{tab:class}
    \end{center}
    \vspace{-8mm}
\end{table}

\subsection{Improvement of object detection performance}
To evaluate the improvement in object detection, the confidence scores and class labels output by the baseline detector are updated iteratively through Algorithm~1.
\begin{algorithm}[htb]
\SetNoFillComment
\DontPrintSemicolon
\SetInd{0.1em}{0.7em}
\SetAlgoLined
$score^{\rm orig.}, label^{\rm orig.}$ $\leftarrow$  output of baseline detector\;
\For{i = 0 {\rm to} 2}  { 
  $score^{\rm calib.}, label^{\rm calib.}$ $\leftarrow$  output of calib. model\;
  \If(\tcp*[f]{update score}) {$i$ = 0 {\rm or} $i$ = 2} {
    $score^{\rm pred}$ $\leftarrow$ $score^{\rm calib.}_{\rm pred\ class}$ $\times$ (1 $-$ $score^{\rm orig.}_{\rm bkg.}$)
    }
    \Else(\tcp*[f]{update label})
     {
        \lIf{$score^{\rm calib.} > 0.98$}  {$label^{\rm pred}$ $\leftarrow$ $label^{\rm calib.}$}
        \lElse {$label^{\rm pred}$ $\leftarrow$ $label^{\rm orig.}$}
     }
   }
\Return{$score^{\rm pred}$, $label^{\rm pred}$ \tcp*[f]{updated values}}
 \caption{Update procedure of label and score}
\end{algorithm}
Here, $label^{\rm orig.}$ and $label^{\rm calib.}$ are the predicted-class labels by the baseline detector and the calibration model, and $score^{\rm orig.}$ and $score^{\rm calib.}$ are the corresponding confidence scores, respectively. 
The quantity of $score^{\rm calib.}$ is calculated by applying a softmax function to the outputs of the calibration model.
The factor of $(1 - score^{\rm orig.}_{\rm bkg.})$, where $score^{\rm orig.}_{\rm bkg.}$ is the baseline confidence score of the bounding box being a background, is intended to suppress drastic change of the predicted score. Table~\ref{tab:objdet} summarizes observed Average Precision (AP) calculated by the COCO API~\cite{cocoapi} for different configurations of the baseline detector. 
 An improvement of 0.8--1.9 in the AP metric by the calibration mode is observed.
\begin{table}[htb]
\tabcolsep = 5.6pt

  \caption{Summary of object detection performance. The naming convention of the object detector follows Detectron2~\cite{wu2019detectron2}. AP is averaged over 10 IoU thresholds between 0.5 and 0.95 with a step of 0.05. AP$^{50(75)}$ is the value of the metic at IoU = 0.50 (0.75). AP$^{s,m,l}$ are AP for object sizes  small (area $<$ 32$^{2}$), medium (32$^{2}$ $<$ area $<$ 96$^{2}$) and large (area $>$ 96$^{2}$), respectively.}

    \begin{center}
  \begin{tabular}{|c|c|c|c|c|c|c|}\hline\hline
 & AP & AP$^{50}$ & AP$^{75}$ & AP$^{s}$ & AP$^{m}$ & AP$^{l}$ \\\hline\hline
R50-C4 & 38.4 & 58.7 & 41.3 & 20.7 & 42.7 & 53.1 \\\hline
+ our model& {\bf 40.3} & {\bf 61.7} & {\bf 43.2} & {\bf 23.2} & {\bf 44.9} & {\bf 54.9} \\\hline\hline
R101-C4 & 41.1& 61.4 & 44.1 & 22.2 & 45.5 & 55.9 \\\hline
+ our model& {\bf 42.8}& {\bf 64.2} & {\bf 45.7} & {\bf 24.7} & {\bf 47.2} & {\bf 58.0} \\\hline\hline
X101-FPN & 43.0& 63.7 & 46.9 & 27.2 & 46.1 & 54.9 \\\hline
+ our model& {\bf 43.8} & {\bf 64.8} & {\bf 47.5}  & {\bf 28.0} & {\bf 46.9} & {\bf 56.2} \\\hline\hline
  \end{tabular}
    \label{tab:objdet}
    \end{center}
        \vspace{-6mm}
\end{table}

\begin{figure}[htb]
\centerline{\includegraphics[width=8.9 cm]{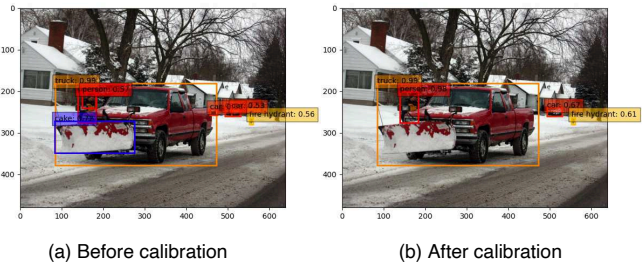}}
  \caption{An example of the detected bounding boxes before the calibration (a) and after the calibration (b). Only the bounding boxes with confidence score greater than 0.5 are shown.}
  \label{fig:example}
\end{figure}

Figure~\ref{fig:example} shows an example of the detected bounding boxes before and after applying the calibration model. The blue bounding box classified as  cake, which is a false positive detection, is observed before the calibration. The calibration model suppresses it properly in this example.

\section{Conclusion}
An improvement of the object detection performance using multi-step ML models is presented in this paper. 
A post-processing model insensitive to the ordering of the input variables, called the calibration model, has been proposed.
The model provides an improvement in AP metric of 0.8--1.9 in our experiments, and can be connected to other object detectors. 
In this study, the baseline object detector and the calibration models were trained independently. Training both of them simultaneously is a future subject to improve accuracy and speed. 

\section{Acknowledgement}
This research was partially supported by Institute of AI and Beyond for the University of Tokyo.

\clearpage


\bibliographystyle{IEEEbib}
\bibliography{refs}

\end{document}